\documentclass[sigconf]{acmart}
\usepackage{graphicx}
\usepackage{hyperref}
\usepackage{listings}
\usepackage{booktabs}
\usepackage{multirow}
\usepackage{amsmath}
\usepackage[most]{tcolorbox}
\usepackage{xcolor}
\usepackage{booktabs} 
\usepackage{amsmath}
\usepackage{amsthm}
\usepackage{multirow,tabularx,ragged2e}
\usepackage{adjustbox}

\setcopyright{acmlicensed}
\copyrightyear{2024}

\acmYear{2024}
\acmDOI{XXXXXXX.XXXXXXX}

\acmConference[SIGSPATIAL '24]{}{October 29 -- November 1,
    2024}{Atlanta, GA}
\acmISBN{978-1-4503-XXXX-X/18/06}

 \def\andreas#1{{\footnotesize{\sc\textcolor{red}{}}}}
 \def\andi#1{{\footnotesize{\sc\textcolor{red}{}}}}
 \def\hossein#1{{\footnotesize{\sc\textcolor{blue}{}}}}
 \def\lance#1{{\footnotesize{\sc\textcolor{blue}{}}}}

\begin{document}

\title{Kinematic Detection of Anomalies in Human Trajectory Data}
\author{Lance Kennedy}
\orcid{}
\affiliation{%
  \institution{Emory University, USA}
  \city{}
  \state{}
  \country{}
}
\email{lance.kennedy@emory.edu}

\author{Andreas Z{\"u}fle}
\orcid{}
\affiliation{%
  \institution{Emory University, USA}
  \city{}
  \state{}
  \country{}
}
\email{azufle@emory.edu}

\renewcommand{\shortauthors}{Kennedy et al.}
\renewcommand{\shortauthors}{Kennedy et al.}
\begin{abstract}
    Historically, much of the research in understanding, modeling, and mining human trajectory data has focused on where an individual stays. Thus, the focus of existing research has been on \emph{where} a user goes. On the other hand, the study of \emph{how} a user moves between locations has great potential for new research opportunities.
Kinematic features describe how an individual moves between locations and can be used for tasks such as identification of individuals or anomaly detection. Unfortunately, data availability and quality challenges make kinematic trajectory mining difficult.
In this paper, we leverage the Geolife dataset of human trajectories to investigate the viability of using kinematic features to identify individuals and detect anomalies. We show that humans have an individual ``kinematic profile'' which can used as a strong signal to identify individual humans. We experimentally show that, for the two use-cases of individual identification and anomaly detection, simple kinematic features fed to standard classification and anomaly detection algorithms significantly improve results. 


\andreas{No References in the Abstract. The Abstract needs to be self-contained. People who read the abstract may not have access to the full paper and the references. The PII claim is a little far fetched! Results would need to be much better to make such claim, and we're not the authorities to decide what is PII.}

\vspace{-0.5cm}
\end{abstract}

\maketitle

\vspace{-0.2cm}
\section{Introduction}
\label{sec:introduction}
Most often, when working with human trajectory data, research has focused on mining and analyzing the locations where a user arrives and stays (or what we may call `staypoints'). Using these staypoints, one can learn about a user's behavior and daily/weekly/monthly patterns. In fact, having just 4 of an individual's staypoints is sufficient to uniquely identify them \cite{de2013unique}. This makes an individual's staypoints Personally Identifiable Information (PII), requiring certain privacy measures to be taken when working with them.
However, staypoints comprise only a part of human trajectory data. We call the movement of users on roads or paths between these staypoints `trips.' A user's trajectory data can be broken down into its trips and staypoints. Very little research has been done on identifying individuals based on their trips, since studies are generally more focused on \emph{where} an individual goes, not how they get there.

That said, if a user's staypoints can be used to uniquely identify them, then it stands to reason that their trips could as well. Consider two individuals who drive in their car from one staypoint to another at the exact same time. Based solely on their staypoints, they would appear the exact same, but in fact they likely \emph{moved} between the staypoints with completely different kinematic behavior. Some individuals drive faster, while some speed up or slow down more suddenly. Each individual should have a \emph{kinematic profile} which is unique to them and how they move from place to place.

These kinematic profiles have the potential to be PII, which would make it important to protect this data and ensure privacy. Additionally, they could be used to identify kinematic anomalies. Consider, for example, if an individual's phone was stolen. The phone, which tracks the GPS location of the user, would know the kinematic profile of how the owner moves. Sensing a kinematic anomaly in the kinematic profile of the thief, it could take security measures to protect the owner's data. Alternatively, a ship may veer off-course to partake in illegal fishing activities, but falsify its trajectory during that time to appear that it stayed on course. The falsified trajectory could be flagged as kinematically different from the ship's normal trajectory.
While this is all idealistic, it is a challenging problem for several reasons. First, it is hard to collect high-quality human trajectory data. For reasons already discussed, this kind of data is protected for privacy \cite{zheng2015trajectory}, so it needs to be collected from anonymous volunteers. One of the most well-known examples is the Geolife GPS trajectory dataset \cite{geolife, zheng2010geolife}, which collected GPS data from 182 users in Beijing China, over a period of several years. Unfortunately, due to the volunteers selectively using the GPS tracking, inconsistent sampling rates, and GPS location errors, it is still difficult to mine accurate kinematic data from this dataset. However, our results show that we can still do so sufficiently to both classify users by their kinematic data, and identify kinematic anomalies. \lance{Need to do more to define the problem}

This paper is organized as follows: In Section~\ref{sec:related_works}, we provide background information and highlight previous results. In Section~\ref{sec:data}, we describe the dataset, before explaining how we mine the kinematic features in Section~\ref{sec:methodology}. In Section~\ref{sec:results}, we present our experimental results. We also provide case studies for users on whom the models worked particularly well and conclude in Section~\ref{sec:conclusion}.

\vspace{-0.4cm}
\section{Related Works}\vspace{-0.1cm}
\label{sec:related_works}
Human trajectory data is spatio-temporal data which records the locations of humans at different timestamps. It has been shown to be PII, 
as it can be used accurately for user classification \cite{de2013unique, seglem2017privacy}. \lance{Do I need more in regards to classification? You could say that anomaly detection is a sub-problem of classification}

Meng et al. \cite{meng2019overview} surveys and categorizes various methods of anomaly detection trajectory data. Lee et al. \cite{lee2007trajectory}, for instance, partitions trajectories into sub-trajectories and focuses on grouping these, rather than looking at them in whole. This serves as the basis for many trajectory anomaly detection algorithms.

Some of the only trajectory anomaly detection algorithms which consider kinematic features concern maritime vessels. Fu et al. \cite{fu2017finding} uses the vessel speed as a feature, but passes that with other features into a semantic trajectory similarity model. Lei \& Mingchao \cite{lei2018distance} also use vessel speed as a feature for a clustering-based approach. Neither of these use kinematics beyond speed and direction. \lance{\cite{machado2019vessel} is in the bib file already but I can't access it for some reason}

\section{Description of Data}\vspace{-0.1cm}
\label{sec:data}
The Geolife GPS trajectory dataset \cite{geolife, zheng2010geolife} was collected from 182 individuals in Beijing, China between 2007 and 2012. It records, for an individual user, a sequence of timestamps, latitudes, longitudes, and other spatial data  such as altitude over some period of time. For our purposes, we do not make use of the other spatial data recorded in the dataset, as the kinematic features we use can all be derived from distance calculations based on latitude-longitude pairs.

While many of the users carried their GPS trackers over long periods of times, about 24\% only did so for less than a week, and only about 42\% of them did so for more than a month. This reduced the amount of users with sufficient, usable data. Furthermore, many of the users exhibit inconsistent sampling rates, where the amount of time between timestamps can vary wildly. \lance{example?} This not only makes it more difficult to mine kinematic features from the users, but along with GPS errors, makes it difficult to determine when a user is at a staypoint, trip, or simply not using the tracker.
Fortunately, the dataset also provides a list of trips, annotated by their \emph{modality}, or method of transportation (e.g. train, car, walk), for 73 of the users. These trips include a start time, end time, and modality for the user's movements. Using these, we no longer need to analyze the raw trajectories to extract trips, although it again reduces the amount of available data.
Therefore, for the following study and experiments, we select GeoLife users having (1) annotated trip modality labels, (2) trips which do not appear to have extreme GPS errors, and (3) at least 30 such trips.

\vspace{-0.35cm}
\section{Methodology}\vspace{-0.1cm}
\label{sec:methodology}
Using the trajectory data, and the start and end times for the trips, we can extract kinematic features per trip. We approximate the speed of the user by calculating the distance between the coordinate pairs at consecutive timestamps, and further approximate the acceleration of the user by taking the differences of consecutive speeds. Of course, neither of these can be perfectly accurate: calculating speed as distance over time assumes that the user was moving in a straight line, and calculating acceleration in this way assumes that changes in speed happen instantaneously at each timestamp. Additionally, as the sampling rate decreases, the more rough these approximations become. \lance{Should I add something here to defend the method?}
%
%
Using the speeds and accelerations, we extract 10 kinematic features per trip, listed in Table~\ref{table:kin-features}, along with their mean and standard deviation in the final dataset. 
It is noteworthy that many of these features are not easily interpretable for humans, but may aid a model in learning about the kinematic profile of a user.
After extracting these features, we notice that many trips include impossible values, such as trips spanning a period of over 100 days, or with speeds of up to 3000000 m/s. Therefore, we reduce the set of trips by removing those which constitute an anomaly for any of our 10 features. These anomalies are determined as follows
: Say \(Q_1, Q_3\) are the values of the data at the 25th and 75th percentile, respectively. Then we let the inter-quartile range \(IQR = Q_3 - Q_1\). We say the anomalies are those which fall outside the range \([Q_1 - 1.5 \cdot IQR, Q_3 + 1.5 \cdot IQR]\). 
Finally, we take only those users which have a sufficient amount of trips after this reduction, a bound which we select as 30. This leaves only 6145 trips over 26 users which are usable for our experiments. Furthermore, the data is considerably imbalanced. The number of trips per user range from 31 to 748. On average, a user has approximately 236 trips, but with great variance; 12 of the users have less than 100 trips.

\begin{table}[t]
\begin{center}
\begin{tabular}{ |c|c|c| }
\hline 
Kinematic Feature & Mean & Std Dev  \\ 
\hline

Trip Duration in $s$& 20144.587 & 23781.262  \\ 

Maximum Speed in $\frac{m}{s}$& 29.006 & 16.737  \\ 

Minimum Speed in $\frac{m}{s}$& 0.0 & 0.0  \\ 

Max. Positive Acceleration in $\frac{m}{s^2}$ & 21.845 & 16.439  \\ 

Min. Negative Acceleration in $\frac{m}{s^2}$& -20.038 & 15.606  \\ 

Mean Speed in $\frac{m}{s}$& 4.095 & 2.361  \\ 

Mean Absolute Accelerations in $\frac{m}{s^2}$ & 0.946 & 0.340  \\ 

Std Dev of Speed in $\frac{m}{s}$ & 3.889 & 1.963  \\ 

Std Dev of Acceleration in $\frac{m}{s^2}$ & 1.978 & 0.942  \\ 

Std Dev of Absolute Accelerations in $\frac{m}{s^2}$& 1.714 & 0.930  \\ 
\hline
\end{tabular}
\caption{Summary of Kinematic Features Used in this Work.\vspace{0cm}}\label{table:kin-features}\vspace{-1.0cm}
\end{center}
\end{table}

\section{Experiments and Results}
\label{sec:results}
We provide two experiments: in Section~\ref{sec:class}, we use the kinematic features to perform user-wise classification of the trips. This is done with a simple decision tree over the kinematic features. In Section~\ref{sec:outlier}, we perform anomaly detection by taking the trips of a single user and adding trips randomly selected from other users as anomalies. This detection is done using the Local Outlier Factor (LOF) algorithm \cite{breunig2000lof}. In both experiments, we provide a case study for one user who provides particularly promising results. The code to generate the data and run these experiments is available at \url{https://github.com/lancek23/trajectory-kinematics}.

\vspace{-0.3cm}
\subsection{Classification}\label{sec:class}\vspace{-0.1cm}

Trips are classified user-wise over 5 stratified folds. The stratification is done to help combat the imbalanced nature of the dataset. This way, an user with as few as 30 trips is guaranteed to have at least 6 trips in each fold of the data, helping the decision tree to train and be tested on a sufficient amount of data per user.

We evaluate the results of classification by the following metrics: accuracy, area under the receiver operating characteristic curve (ROC-AUC), and F1 score. The ROC-AUC helps us ensure that the model works better than a random classifier. The F1 score uses ``macro" aggregation, taking the unweighted mean of scores across the different class labels, and serves as a conservative metric for class-imbalanced problems which evaluates the model based on precision and recall.

We provide two random classifiers against which we measure our results. The first classifier uses a ``weighted random guess" in which it predicts a label with a probability equal to the proportion of that label in the dataset. The second classifier uses a ``true random guess," and predicts each label with equal probability. The results can be seen in Table~\ref{table:class-results}, aggregated over the 5 folds. The ROC-AUC shows that our model is superior to any random classifier, while we also show significant improvements in terms of accuracy and F1 score, correctly classifying around 30\% of all trips in the dataset.

\begin{table}[t]
\begin{center}
\begin{tabular}{ |c|c|c|c| }
\hline 
Model & Accuracy & ROC-AUC  & F1 score \\ 
\hline

Decision Tree & $\mathbf{.303 \pm .008}$ & $\mathbf{.589 \pm .004}$ & $\mathbf{.208 \pm .006}$ \\ 

Weighted Guess & $.079 \pm .005$ & $.500$ & $.041 \pm .005$  \\ 

Random Guess & $.035 \pm .003$ & $.500$ & $.026 \pm .002$  \\ 
\hline
\end{tabular}
\captionof{table}{Results compared to random classifiers}\label{table:class-results}
\end{center}
\vspace{-1.1cm}
\end{table}

\begin{figure}
    \centering
    \includegraphics[width=1\columnwidth, height=6.8cm]{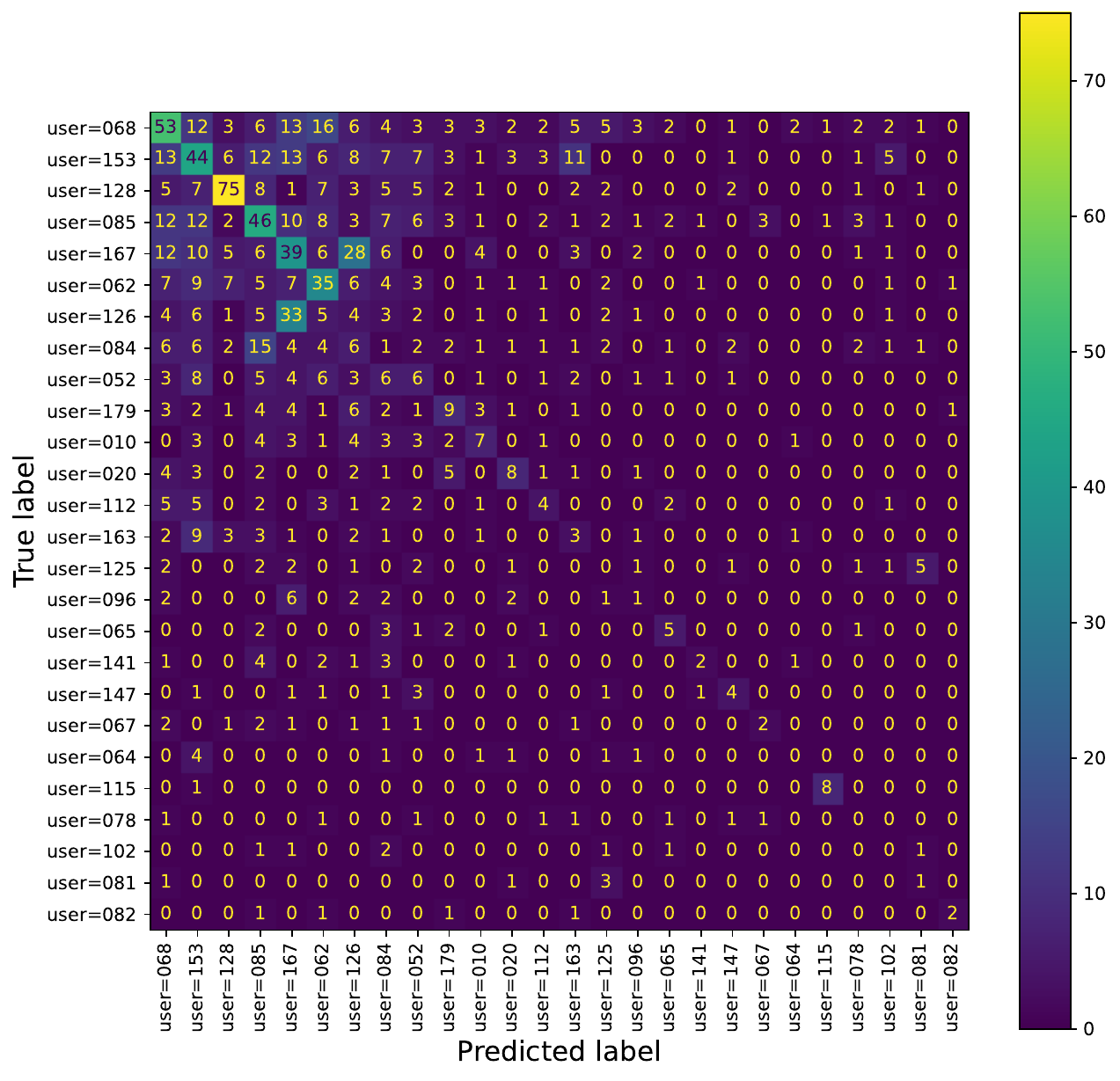}
    \vspace{-0.9cm}
    \caption{Confusion matrix for the decision tree model. Users are sorted by their number of trips.}
    \label{fig:cm}
    \vspace{-0.6cm}
\end{figure}

We now provide an example confusion matrix in Figure~\ref{fig:cm}. The user labels are organized in this matrix from most to least trips, for convenience. We can observe that the decision tree lacks the bias that the weighted random guess will have towards those users with more trips, and is still able to correctly predict some of the trips for those users with relatively few trips. We observe that the kinematics based user identification is substantially better than random guessing. We also observe that for some users, the classifcation is very accurate. For example, for User 068, we observe a Recall of $36\%$ and a Precision of $38\%$. For User 128, we observe a Recall of $69\%$ and a Precision of of $71\%$. This are substantially higher than the Precision/Recall values of $1/26\sim 4\%$ that we would expect using random guessing. We also observe that the model confuses some users. For example, User 126 is frequently misclassified as User 167, having a Recall of only $\%$ and a Precision of only $5\%$. What makes some users easier to detect kinematically than others?

\vspace{-0.2cm}
\subsubsection{Case Study}
\begin{figure}
    \centering
    \includegraphics[width=1\columnwidth, height=5cm]{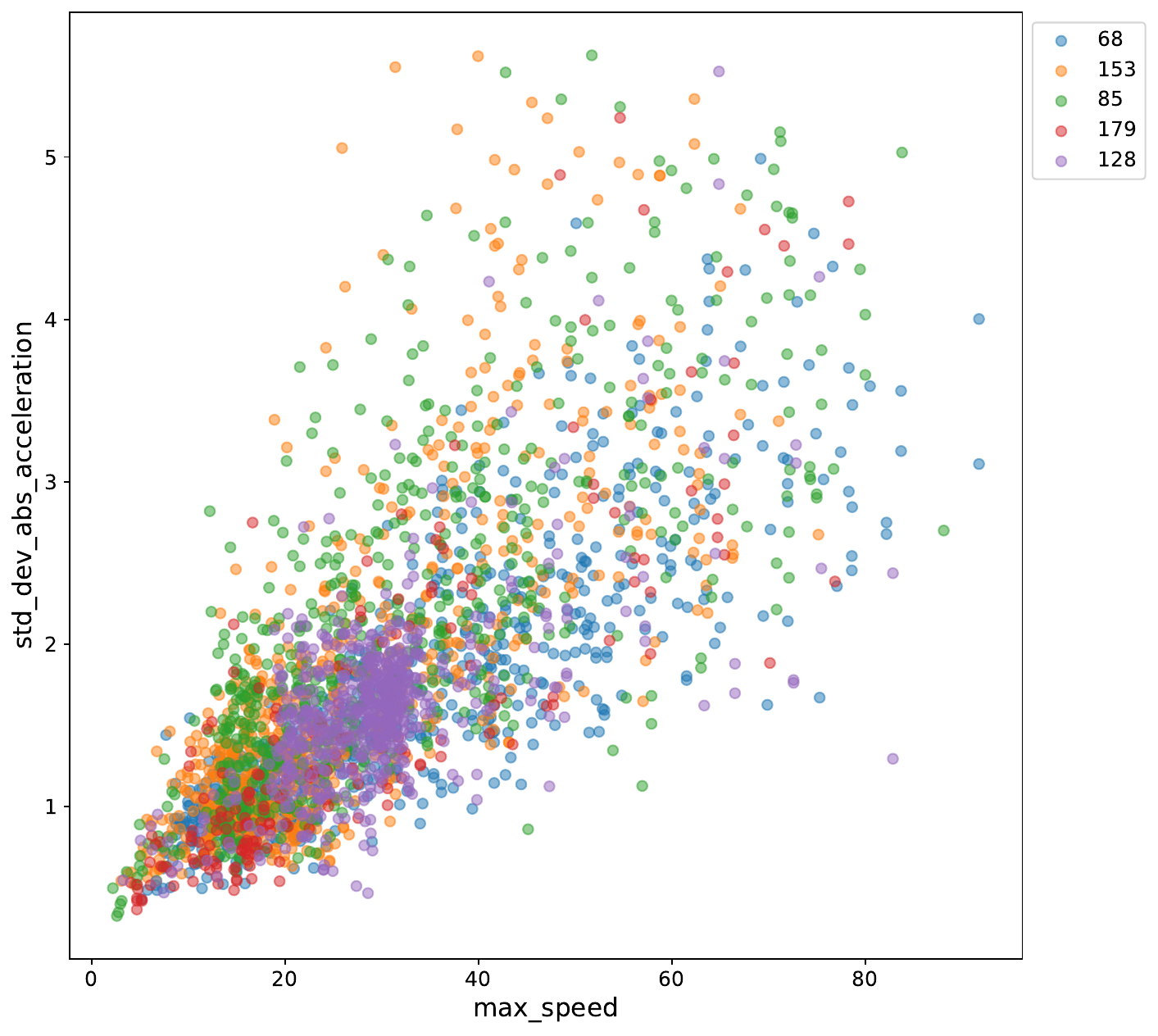}
    \vspace{-0.8cm}
    \caption{A plot of kinematic features of 5 users. Note the dense purple cluster formed by one of the users.}
    \label{fig:trips}
    \vspace{-0.9cm}
\end{figure}

\begin{figure}
    \centering
    \includegraphics[width=1\columnwidth, height=5cm]{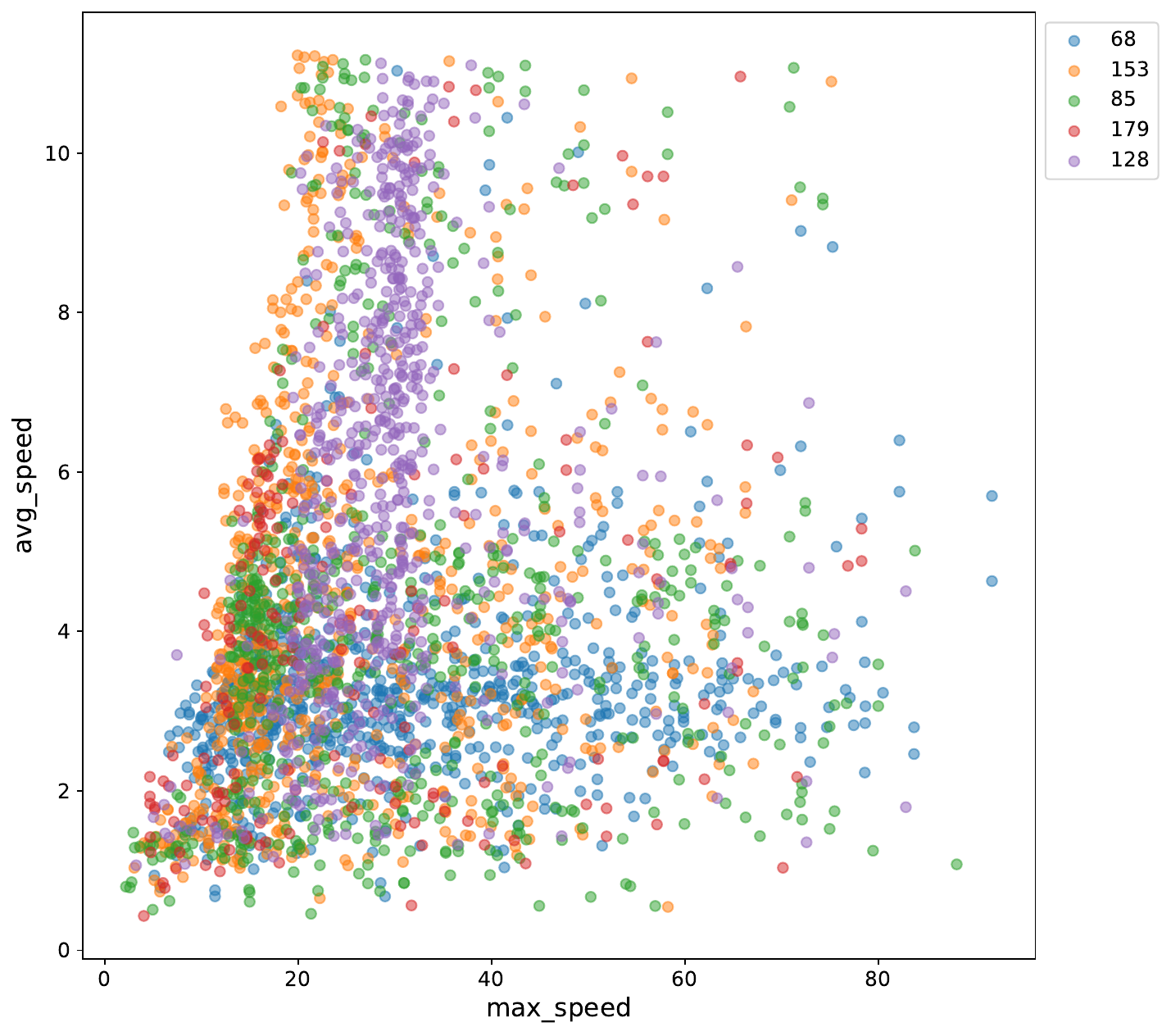}
    \vspace{-0.9cm}
    \caption{A plot of kinematic features of 5 users. Note the consistency of max speed for the user plotted in purple.}
    \label{fig:speed}
    \vspace{-0.6cm}
\end{figure}

To investigate this question, we provide a case study for a user we call ``User 128." User 128 has noticeable kinematic patterns, particularly in their maximum speed and in their standard deviation of absolute accelerations per trip. The trips of User 128 make up the dense, purple cluster seen in Figure~\ref{fig:trips}.
We can again observe the kinematic profile of User 128 in Figure~\ref{fig:speed}, which plots the maximum speed vs the average speed of the trips for a few users. We see that despite changes in average speed, the maximum speed achieved by User 128 appears consistent. Alternatively, some of the other users plots look nearly uniform, with less apparent trends. \lance{I should probably pick worse looking users for this...} In reality, it is likely that the other users achieve more consistent maximum speeds during different trips, but data collection errors and low sampling frequencies lead to issues in capturing this.

User 128 was consistently easy for the decision tree to classify. In Figure~\ref{fig:cm}, we see that more of the true positives we find are from User 128 than any other, despite having only the third-most trips in the dataset. While this is partially explainable from decision trees having a natural bias towards labels with more observations in a dataset, User 128 is still much easier to identify than the other users with a large number of trips. This is because users 128's trips are more well-clustered than the trips of other users. \lance{Could talk about excellent recall vs okay precision, because the large number of trips + good clustering leads to the model wanting to assign other similar trips to this label}\andi{Added that in the previous section.}
User 128 provides an example of what is possible in using kinematics to discriminate between human trajectories. With more accurate datasets in the future, we hope to show that the results for User 128 can be replicated on a greater number of human trajectories. 

\subsection{Anomaly Detection}\label{sec:outlier}

To perform anomaly detection, we first inject anomalous mobility by swapping user IDs of trips, thus assigning a trip of one user to another user.
For each user, we randomly select trips from the other users up to roughly 3\% of the number of trips the original user has in the dataset. This way, the trips of the original user should have a consistent kinematic profile based on how the user moves, while the other trips form a minority anomaly class which should be kinematically different from those in the normal class. These trips together form our new dataset.

Using the LOF algorithm, we assign an anomaly score to each trip based on their local density \cite{breunig2000lof}. We expect that the trips of the normal user will be kinematically similar and will form a dense cluster, while the anomaly trips will be far from this cluster with few nearby neighbors. 
Furthermore, the different modalities (walk, car, etc) of the normal user should each form their own cluster, with the inserted trips apart from each of them. Sorting the trips by their anomaly score, we find the area under the precision-recall curve (PR-AUC), which is a useful indicator for binary data.

We allow each of our 26 users to serve as the ``normal" user for 10 trials, randomly adding anomalies for each trial as described, for 260 total trials. We also run the same experiments but assign anomaly scores using a random classifier rather than the LOF algorithm as a baseline. The results are summarized in Table~\ref{table:outlier-results}.

\begin{table}[t]
\begin{center}
\begin{tabular}{ |c|c|c| }
\hline 
& LOF & Random Classifier \\ 
\hline

Mean & 0.087 & 0.046 \\ 

Std & 0.164 & 0.057 \\ 

Min & 0.014 & 0.011 \\ 

Median & 0.042 & 0.030 \\ 

Max & 1.000 & 0.518 \\ 
\hline
\end{tabular}
\captionof{table}{PR-AUC summary statistics}\label{table:outlier-results}\vspace{-0.9cm}
\end{center}
\end{table}

These results are, unfortunately, not as good as we had hoped. We're still able to outperform a random baseline, which uniformly at random selects 3\% of trips as anomalies. But the results of the Local Outlier Factor having a mean AUC-PR of 8.7\% is not practically useful. This is surprising, as we that that the decision tree classifier was able to accurately identify the user that generated a trip. Yet, the Local Outlier Factor struggles to detect an agent changing their identity. A problem might be the large number of kinematic features that we consider, which may cause the LOF, which computes distances between multi-dimensional points, to run into the Curse of Dimensionality~\cite{indyk1998approximate}. It may be possible to automatically find the right subspaces to detect outliers in as proposed in~\cite{kriegel2009outlier}, but we leave this study for future work. 
Additionally, we believe that much of this is due to the previously mentioned issues with data collection. If there are errors in mining the kinematic features of a user, then it may be impossible to find the true patterns for that user and build an accurate kinematic profile. However, there are some users which we are able to perform better on than others. We believe that these users are not necessarily \emph{easier} to discriminate from others, but rather that these users have more accurate data, allowing us to build accurate kinematic profiles of them. We provide a case study of one such user in section~\ref{sec:outlier-case-study}.

\begin{figure}
    \centering
    \includegraphics[width=1\columnwidth, height=6cm]{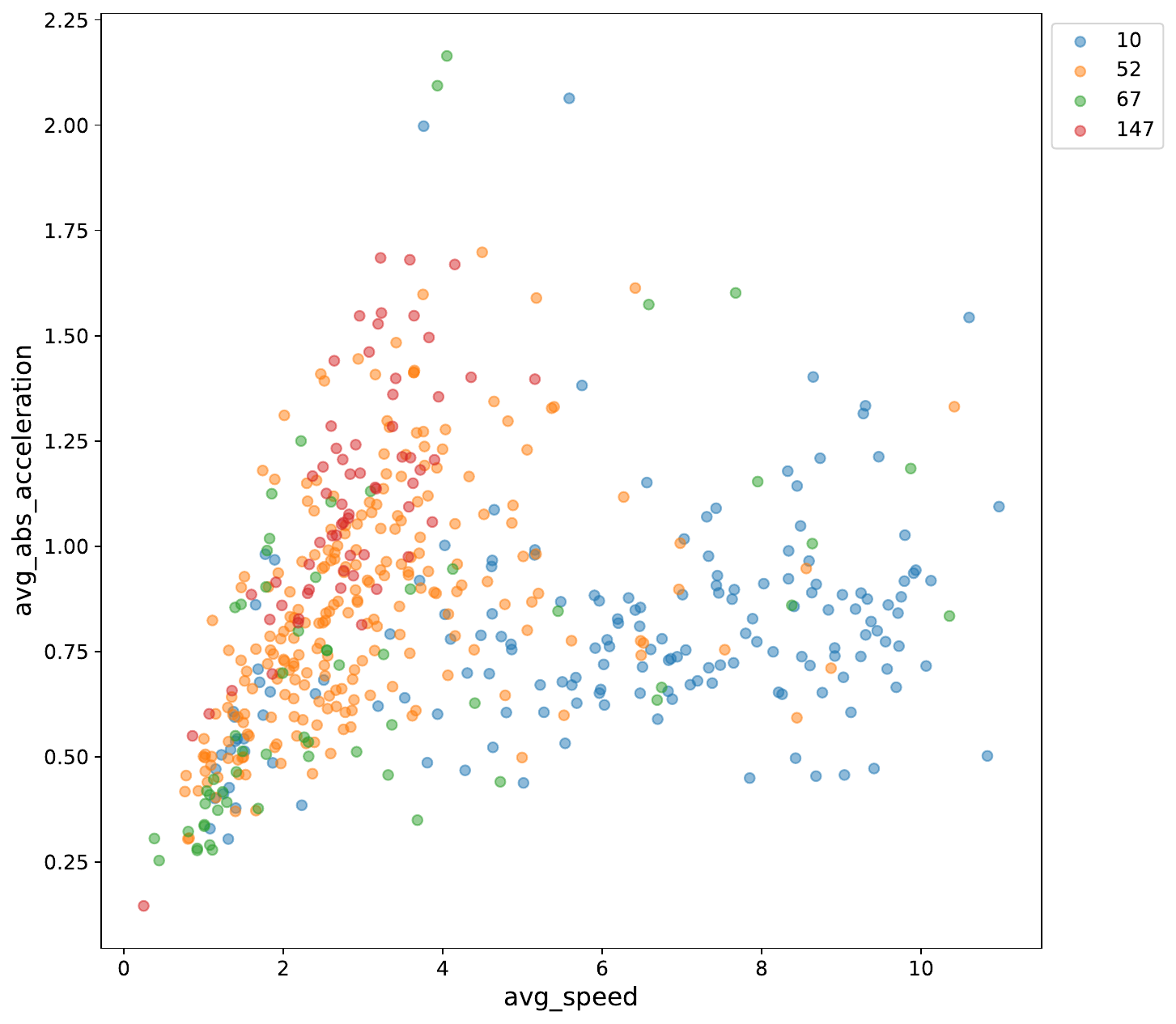}
    \vspace{-0.6cm}
    \caption{A plot of kinematic features of 4 users. Note the consistency of average speed for the user plotted in red.}
    \label{fig:agent147}
    \vspace{-0.8cm}
\end{figure}

\subsubsection{Case Study}\label{sec:outlier-case-study}

The user we call ``User 147" consistently performed well in anomaly detection. Figure~\ref{fig:agent147} shows this user, along with some others, plotted by two kinematic features. We notice that User 147, plotted in red, is very consistent in its average speed compared to the other users. This explains why User 147 is so easy to find anomalies with: any trip with an average speed outside of its typical range can be marked as an anomaly. In the classification problem, User 147 wasn't as easy to identify due to so many other trips having similar average speeds, but when it represents the majority ``normal" class, this is no longer an issue. In fact, in the 10 trials where User 147 was ``normal," the LOF algorithm had an average PR-AUC of 0.29, even achieving 
0.792 in its best trial. \lance{Include figure?} User 147 demonstrates an important principle: when the data provides a consistent kinematic profile, the anomaly detection problem becomes easy.

\vspace{-0.2cm}
\section{Conclusion and Future Work}\vspace{-0.1cm}
\label{sec:conclusion}
This paper provides an early look into the potential of kinematic data mined from human trajectory data. Using relatively simple models, we are able to classify trajectories by user, and identify anomalous trips from a user of interest.

This has important implications in privacy, as kinematic data may be PII, and therefore important to keep under privacy restrictions. It also has practical uses in identifying unusual or falsified trajectories. These may signify a phone being stolen as the kinematic profile of the user changes, or a ship falsifying their trajectory to cover up illegal behavior.
With more accurate datasets in the future, the ability to perform classification and anomaly detection tasks using kinematic data could be even greater. In addition, more complex models may be able to detect new patterns in the kinematic profiles, yielding improved performance.

 \vspace{-0.2cm}
\section{Acknowledgements}\vspace{-0.1cm}
 Supported by the Intelligence Advanced Research Projects Activity (IARPA) via Department of Interior/ Interior Business Center (DOI/IBC) contract number 140D0423C0025. The U.S. Government is authorized to reproduce and distribute reprints for Governmental purposes notwithstanding any copyright annotation thereon. Disclaimer: The views and conclusions contained herein are those of the authors and should not be interpreted as necessarily representing the official policies or endorsements, either expressed or implied, of IARPA, DOI/IBC, or the U.S. Government.
 \vspace{-0.35cm}

\bibliographystyle{ACM-Reference-Format}
\bibliography{refs/main}

\end{document}